\documentclass[final]{cvpr}

\usepackage{times}
\usepackage{epsfig}
\usepackage{graphicx}
\usepackage{amsmath}
\usepackage{amssymb}

\usepackage{amsthm}
\newtheorem{theorem}{Theorem}

\newtheorem{claim}{Claim}

\usepackage{booktabs}
\usepackage{subfigure}
\usepackage{caption}
\usepackage{color}
\usepackage{multirow}
\usepackage[pagebackref=true,breaklinks=true,colorlinks,bookmarks=false]{hyperref}

\usepackage[marginal]{footmisc}

\begin{document}

%%%%%%%%% TITLE
% \title{SPSL: Spatial-Phase Shallow Learning for Face Forgery Detection}
\title{Spatial-Phase Shallow Learning: Rethinking Face Forgery Detection \\ in Frequency Domain}

% \author{Honggu Liu\\
% University of Science and Technology of China\\
% Institution1 address\\
% {\tt\small lhg9754@mail.ustc.edu.cn}
% % For a paper whose authors are all at the same institution,
% % omit the following lines up until the closing ``}''.
% % Additional authors and addresses can be added with ``\and'',
% % just like the second author.
% % To save space, use either the email address or home page, not both
% \and
% Second Author\\
% Institution2\\
% First line of institution2 address\\
% {\tt\small secondauthor@i2.org}
% \and
% Third Author\\
% Institution2\\
% First line of institution2 address\\
% {\tt\small secondauthor@i2.org}
% }

\author{Honggu Liu\textsuperscript{\rm 1}\textsuperscript{$\ast$}\qquad Xiaodan Li\textsuperscript{\rm 2}\qquad Wenbo Zhou\textsuperscript{\rm 1}\textsuperscript{$\dagger$}\qquad Yuefeng Chen\textsuperscript{\rm 2}\\
Yuan He\textsuperscript{\rm 2}\qquad Hui Xue\textsuperscript{\rm 2}\qquad Weiming Zhang\textsuperscript{\rm 1}\textsuperscript{$\dagger$}\qquad Nenghai Yu\textsuperscript{\rm 1}\\ \textsuperscript{\rm 1}University of Science and Technology of China \quad \textsuperscript{\rm 2}Alibaba Group\\
{\tt\small lhg9754@mail.ustc.edu.cn,}\quad {\tt\small {\{welbeckz, zhangwm, ynh\}@ustc.edu.cn}} \\ 
{\tt\small {\{fiona.lxd, yuefeng.chenyf, heyuan.hy, hui.xueh\}@alibaba-inc.com}}}

\maketitle

\footnote{\textsuperscript{$\ast$}\  Work done during the internship of Honggu Liu at Alibaba Group.}
\footnote{\textsuperscript{$\dagger$}\  Corresponding Author.}

%%%%%%%%% ABSTRACT
\begin{abstract}
The remarkable success in face forgery techniques has received considerable attention in computer vision due to security concerns. We observe that up-sampling is a necessary step of most face forgery techniques, and cumulative up-sampling will result in obvious changes in the frequency domain, especially in the phase spectrum. According to the property of natural images, the phase spectrum preserves abundant frequency components that provide extra information and complement the loss of the amplitude spectrum.
% For this reason: To this end, we present a novel \textbf{S}patial-\textbf{P}hase \textbf{S}hallow \textbf{L}earning~(\textbf{SPSL}) based on the phase spectrum for face forgery detection. In this paper, the SPSL method combines spatial image and phase spectrum to capture the up-sampling artifacts of face forgery which improves the transferability of our proposed method.
To this end, we present a novel \textbf{S}patial-\textbf{P}hase \textbf{S}hallow \textbf{L}earning~(\textbf{SPSL}) method, which combines spatial image and phase spectrum to capture the up-sampling artifacts of face forgery to improve the transferability, for face forgery detection. And we also theoretically analyze the validity of utilizing the phase spectrum.
Moreover, we notice that local texture information is more crucial than high-level semantic information for the face forgery detection task. So we reduce the receptive fields by shallowing the network to suppress high-level features and focus on the local region. Extensive experiments show that SPSL can achieve the state-of-the-art performance on cross-datasets evaluation as well as multi-class classification and obtain comparable results on single dataset evaluation. 
% \xx{We also theoretically analyze the validity of the proposed SPSL method. Delete or change the place.}

\end{abstract}

\section{Introduction}
Benefiting from the tremendous success of generative techniques, such as Variational Autoencoders~(VAE)~\cite{pu2016variational} and Generative Adversarial Networks~(GANs)~\cite{goodfellow2014generative}, face forgery has become an emerging hot research topic in very recent years. The face forgery techniques are able to synthesize realistic faces that are indistinguishable for human eyes.
% Meanwhile, it's a hard job for people to identify whether it is a synthetic image or a real one.
%  With the advanced techniques in image synthesis, it is almost impossible even for humans to distinguish the real from fake. 
However, these forgery techniques are likely to be abused for malicious purposes, causing serious security and ethical issues (\eg celebrity pornography and political persecution). Therefore, it is of paramount importance to develop more general and practical methods for face forgery detection.

\begin{figure}[t]
\setlength{\belowcaptionskip}{-0.4cm}
\centering
\includegraphics[scale=0.55]{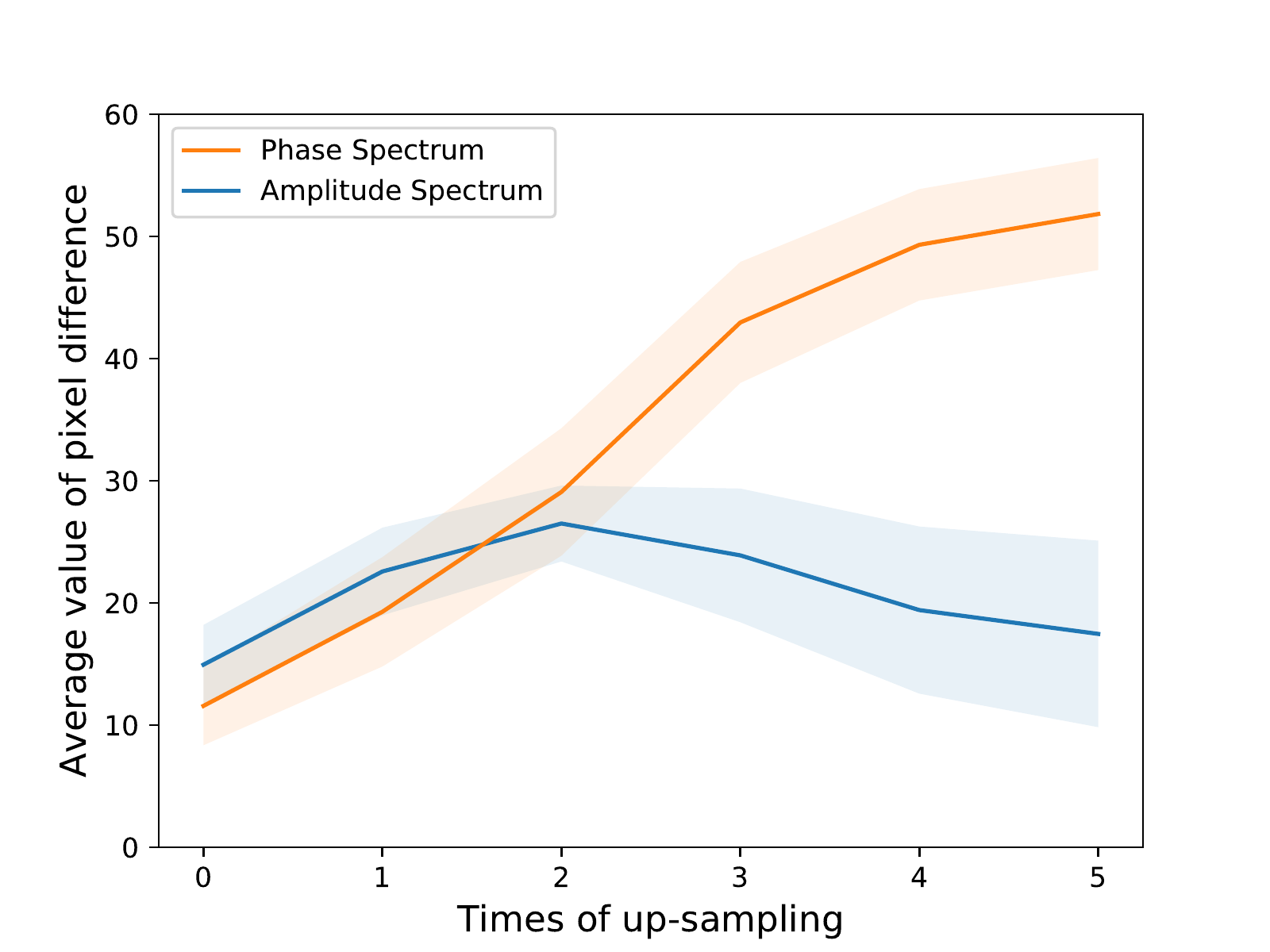}
\caption{\textbf{The variation analysis in the frequency domain}. The curve shows the average value of pixel difference (mean and variance) of the phase spectrum between 1000 origin and up-sampling samples increase dramatically with the increase of the number of up-sampling.}
% \xx{Explain L2 norm}}
\label{fig:amp_phs}
\end{figure}

To alleviate the risks brought by malicious usage of face forgery, various methods~\cite{zhou2017two,li2018exposing,afchar2018mesonet,yang2019exposing, agarwal2019protecting,sabir2019recurrent,rossler2019faceforensics++,durall2019unmasking,li2020face,10.1007/978-3-030-58610-2_6,masi2020two,dang2020detection,li2020sharp} have been proposed.
Most of these methods detect face forgery in a supervised fashion with prior knowledge of face manipulation methods~\cite{afchar2018mesonet,Chollet_2017_CVPR,li2020sharp}. Under this setting, these approaches achieve excellent performance on some public datasets \cite{yang2019exposing,korshunov2018deepfakes,rossler2019faceforensics++,li2020celeb, wang2020cnn}. However, these detection methods tend to suffer from overfitting thus their effectiveness is limited to the datasets which they are specifically trained on. Moreover, in real-world detection, it is inevitable to face a source/target mismatch problem when handling unseen samples, which is extremely challenging for deepfake detection task.
% it is hard to trace the precise data sources when we face unknown samples in a practical scene. 
Therefore, it is necessary to enhance the generalization and transferability of forgery detection techniques. Recently, some approaches~\cite{du2019towards,li2020face,masi2020two,xuan2019generalization} have made attempts to improve the transferability, there are still deficiencies.
% \textcolor{blue}{on such as performance and universality}. 
For example, two-branch~\cite{masi2020two} achieves the state-of-the-art performance of cross-dataset detection at the expense of frame-level detection accuracy. Face X-ray~\cite{li2020face} starts from a novel perspective that aims at detecting the blending boundary artifacts and obtains perfect performances in detecting unseen forgery method for raw videos, which significantly improves transferability of different manipulation methods~\cite{DeepFakes,thies2016face2face, FaceSwap, thies2019deferred}.
% in a forged image left by rendering the altered face into an existing background image.
However, since only focus on the information extracted from the spatial domain, it is easily to be influenced by video compression. Thus, we need to concern about the more common artifacts in the generation of forgery images from various domains.
\begin{figure}[t]
\setlength{\belowcaptionskip}{-0.4cm}
\centering
\includegraphics[scale=0.35]{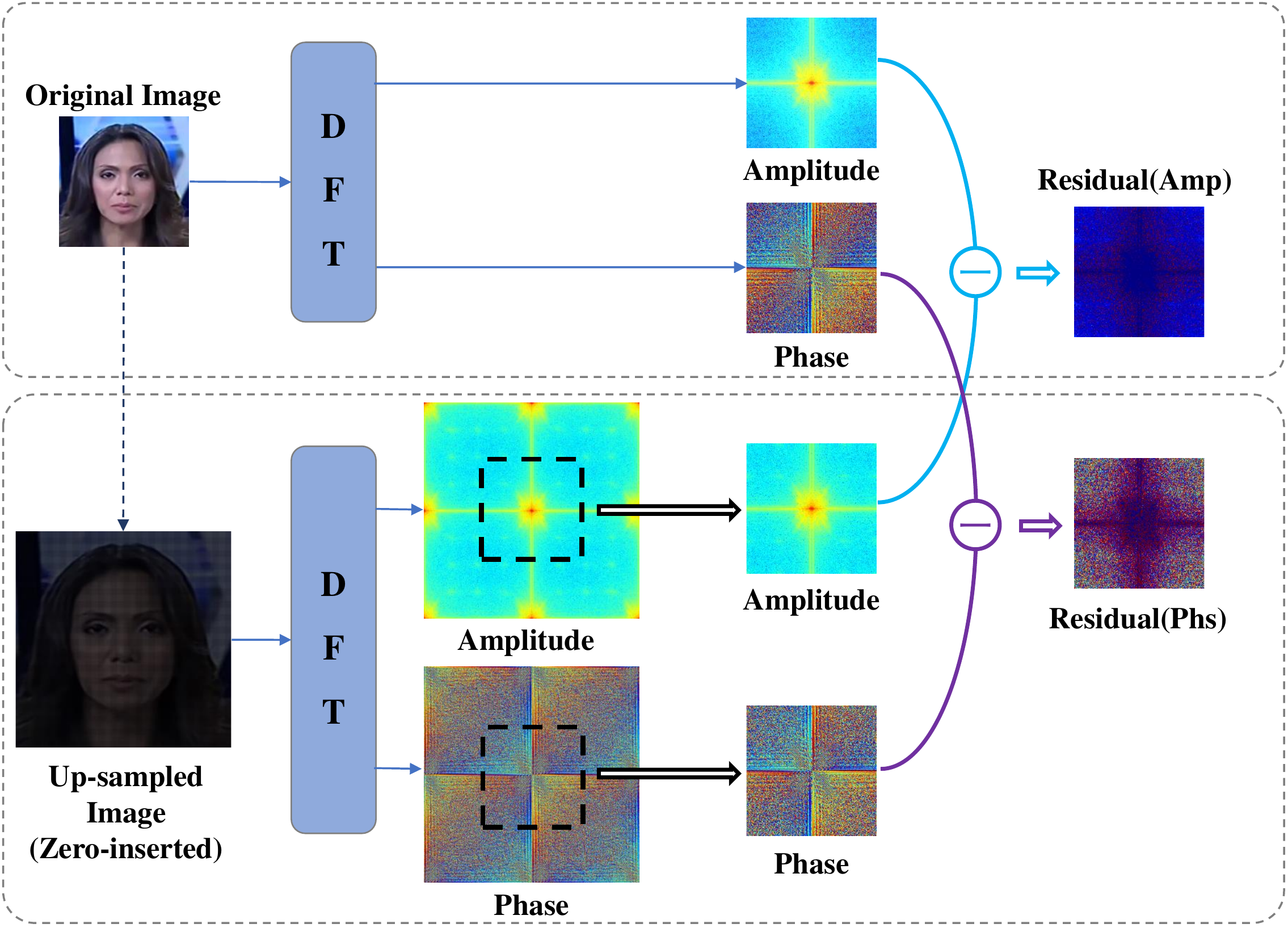}
\caption{The frequency domain analysis of the origin and up-sampling face. The residual images show that the differences of phase spectrum between origin and up-sampling are bigger than the amplitude spectrum with the up-sampling. Best viewed in color. (Darker color indicates smaller pixel values).}
\label{fig:upsample}
\end{figure}

We obverse that \textbf{up-sampling} is a non-negligible step in generative models(\eg VAE~\cite{pu2016variational}, GANs~\cite{goodfellow2014generative}), from which the generated part are then used to synthesize fake faces.
This operation usually leaves a trace in the frequency domain, which provides cues for separating synthesized faces from real ones. 
Though~\cite{durall2020watch} also tried to detect these artifacts with the amplitude spectrum, the performance is limited due to the information loss.

To this end, we propose Spatial-Phase Shallow Learning (SPSL) for face forgery detection, which leverages the phase spectrum for detecting the common artifacts. The pivotal thought is that the phase spectrum is more hypersensitive to up-sampling than the amplitude spectrum. As shown in Figure~\ref{fig:amp_phs}, with more times of up-sampling operations are performed, the average pixel differences of the phase spectrum get much greater than that of the amplitude spectrum. A visualization comparison can also be found in Figure~\ref{fig:upsample}.

Moreover, \cite{chai2020makes} used the patch-based classification to generalize the image forensics. Thus, we hold the opinion that local texture information is more important than high-level semantic information even high-level semantic information should be suppressed to a certain extent in the specific task of forged face detection.
% \yf{Moreover, to capture artifacts, local texture feature is more important than high-level semantic information.}
For this purpose, SPSL drops many convolutional layers to reduce the receptive field~\cite{araujo2019computing} and forces CNNs to pay more attention to local regions which are abundant in textures and lack high-level semantic information.

As a result, SPSL focuses on the common step in the forged faces generation and pays more attention to textures leading to a performance improvement of the cross-datasets evaluation. At the same time, the performance on multi-class classification also improves due to the specific traces of different categories manipulation are left in the phase spectrum. 

Extensive experiments demonstrate that SPSL significantly improved the transferability and achieved the state-of-the-art over cross-dataset evaluation. 
% \textcolor{blue}{We also show that our framework remarkably improves the accuracy of multi-class classification, as well as the ability to widen the distance of four face manipulation algorithms in FaceForensics++~\cite{rossler2019faceforensics++} in feature space.}
% , as shown in Figure~\ref. Meanwhile, we show the efficiency improvement compared to the basic backbone network.

The major contributions in this paper are summarized as follows:
\begin{itemize}
	\item We firstly leverage the phase spectrum to detect forged face images and demonstrate that CNNs can capture extra implicit features of the phase spectrum which are beneficial to face forgery detection with precise mathematical derivation.
% 	proposed a ingenious method of learning frequency information with CNNs.
% 	\item [-] We demonstrate that CNNs can capture extra implicit features of phase spectrum which are benefit to face forgery detection with precise mathematical derivation (Claim~\ref{clm:cnns get more}).
	\item Aiming at the specific problem of forged face detection, we assume that high-level semantic information should be appropriately suppressed. And we experimentally validate the hypothesis by decreasing the receptive field of CNNs with shallow network learning.
	\item We verify that our approach achieves the state-of-the-art performance of forged face detection over cross-dataset evaluation.
% 	\yf{and make a remarkable improvement of multi-class face manipulation detection}.
\end{itemize}

\section{Related work}
Since face manipulation is a classical research topic~\cite{thies2016face2face,petrov2020deepfacelab, nirkin2019fsgan, korshunova2017fast, karras2019style,thies2019deferred} in computer vision, verifying its authenticity is not a new problem.
% Image forgery, face manipulation in particular, has been widely applied for a long time, and thus verifying image authenticity is not a new problem.
However, recent remarkable successes of deep learning make face manipulation easier and more realistic which poses a significant challenge of forged face detection. In this section, we briefly review the current face forgery detection methods that are representative and related to our work.
\subsection{Spatial-based Face forgery Detection}
With the development of face forgery, a wide variety of methods have been proposed to detect forged face. The majority of them exploit artifacts based on the spatial domain, especially in RGB. Some methods for deepfake detection focus on hand-crafted facial features from the video, such as eye blinking~\cite{li2018ictu}, inconsistent head poses~\cite{yang2019exposing}, facial expression change~\cite{agarwal2019protecting}. Recent methods~\cite{afchar2018mesonet,nguyen2019capsule,rossler2019faceforensics++} capture high-level features from the spatial domain by using deep neural networks and show impressive performance.
% MesoNet~\cite{afchar2018mesonet} is a CNN-based network and focuses on the mesoscopic properties of forged images. 
Nguyen~\etal proposed a method~\cite{nguyen2019capsule} which leveraging capsule network~\cite{sabour2017dynamic} to detect face manipulation. Rossler~\etal~\cite{rossler2019faceforensics++} show the best performance on many kinds of forgery algorithms with the efficient XceptionNet~\cite{Chollet_2017_CVPR} at that time. Face X-ray~\cite{li2020face} mainly focuses on the blending step which exists in most face forgery and thus achieved state-of-the-art performance on transferability in raw videos. However, it still has some limitations that the performance of Face X-ray will sharply drop when encounter low-resolution images, and it may not work with entirely synthesized images.
Almost all of these CNN-based methods only use spatial domain information and therefore the performance is quite sensitive to the quality or data distribution of datasets. In our work, we combine the spatial domain with the frequency domain to take advantage of both.
\subsection{Frequency-based Face forgery Detection}
Besides focusing on the spatial domain, some methods pay attention to the frequency domain for capturing artifacts of the forgery. In fact, frequency analysis is a common and important way in digital image processing and has been widely applied to various tasks in computer vision~\cite{wang2020high,xue2020faster,stuchi2017improving,kim2012face}. Most of them use either Discrete Fourier Transform (DFT) or Wavelet Transform (WT), or Discrete Cosine Transform (DCT) to convert the spatial image to the frequency domain. Durall~\etal~\cite{durall2019unmasking} first proposed that averaging the amplitude of each frequency band with DFT can mine abnormal information of forgery in face manipulation detection. $\text{F}^3\text{-Net}$~\cite{10.1007/978-3-030-58610-2_6} extracted frequency-domain information using DCT and analyzed the statistic features for face forgery detection. $\text{F}^3\text{-Net}$ achieved state-of-the-art performance on highly compressed videos, but the performance on cross-dataset evaluation drops greatly. Masi~\etal~\cite{masi2020two} leverage a Laplacian of Gaussian (LoG) to make frequency enhancement for purpose of suppressing the image content present in the low-level feature maps. However, most of these related works mainly depend on low-level statistical features rather than high-level features extracted by CNNs, and therefore the frequency information was inadequately utilized. In our work, given the powerful capabilities of feature extraction of CNNs, we make use of the phase spectrum in DFT and explicitly prove the validity with theory analysis while we integrate it into the whole learning process of CNNs.

\begin{figure}[t]
\setlength{\belowcaptionskip}{-0.4cm}
\centering
\includegraphics[scale=0.55]{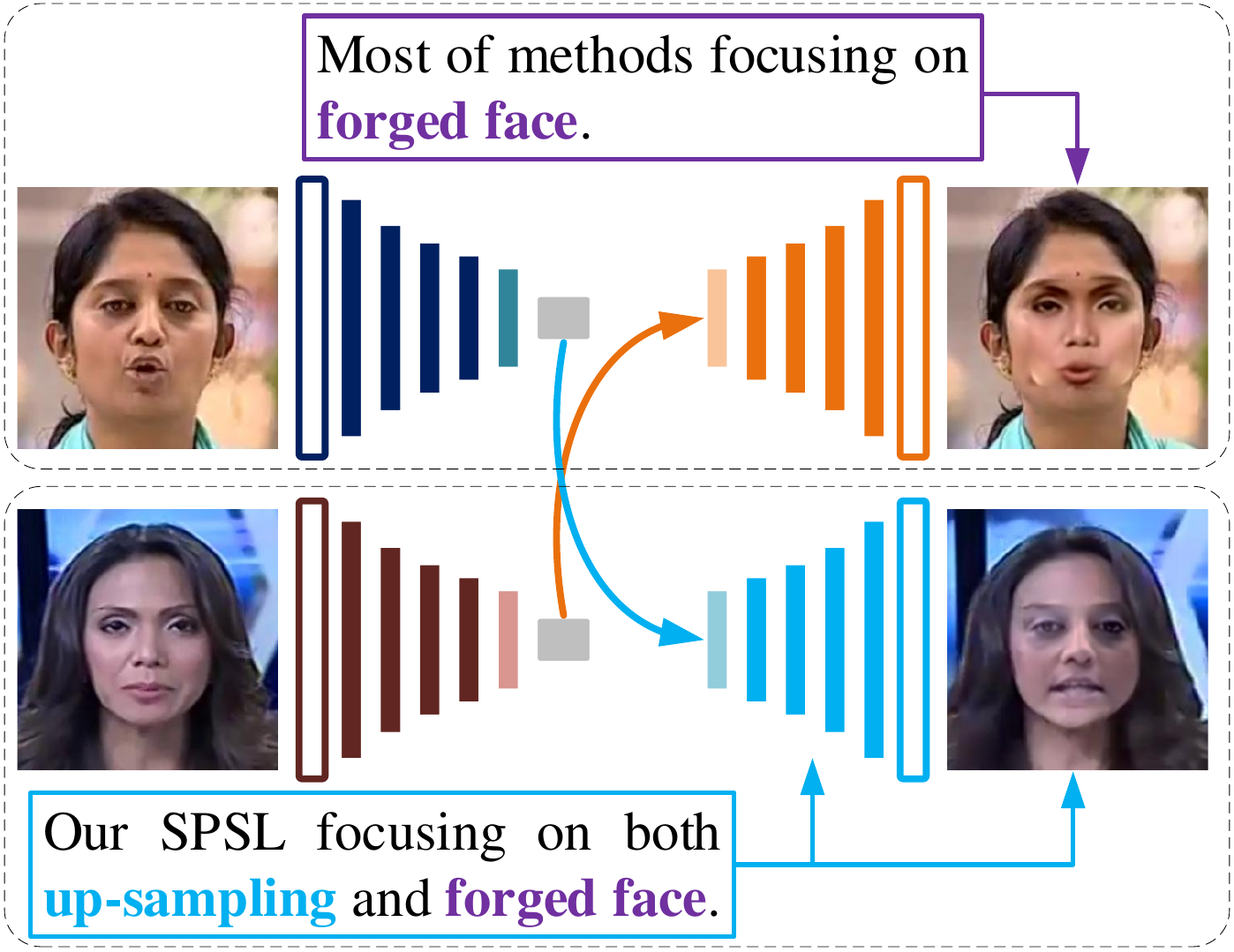}
\caption{Overview of typical face manipulation pipeline. Most of the previous works only focus on forged faces, while we focus on both up-sampling and forged faces.}
\label{fig:motivation}
\end{figure}

\section{SPSL for Face Forgery Detection}
% In this section, we firstly describe the motivation of our method in Section~\ref{sec:motivation} which is the first utilization of phase information to detect forged face images. Then we emphatically reveal the advantages of phase spectrum in deep learning based detection methods in Section~\ref{sec:derivation_1} and Section~\ref{sec:derivation_2}. In Section~\ref{sec:derivation_3}, we proves how the phase spectrum helps deep learning based face forgery detection. Finally, we briefly describe the way of suppressing high-level semantic information and focusing on local textures information and introduce the way of shallowing networks in Section~\ref{sec:shallow}.
In this section, we start by introducing the key observation of face forgery generation. Then we propose spatial-phase shallow learning to detect the observed common artifacts for face forgery detection. Finally, Making the network shallow can focus on the local region to further boost the improvement of transferability.

\subsection{Motivation}~\label{sec:motivation}
% Up-sampling is the vital step of forged face generation with whether AutoEncoders or GANs, and therefore results in the emergency of new frequency components with almost all face forgery techniques. Thus, these new frequency components will bring out some common artifacts in frequency domain which is our main focus point. Former studies usually just analyze the statistics of amplitude spectrum, which completely discard the phase spectrum, and hence the information available is very limited. Thus, how to maximize the use of frequency-domain information with CNNs in face forgery detection is worthy of more careful study. 
As shown in Figure~\ref{fig:motivation}, a typical facial manipulation method consists of three stages~\cite{DeepFakes}: 1) encoding source face; 2) swapping face in latent space; 3) decoding target face.

Up-sampling is a vital step for decoding the target face, based on either AutoEncoders~\cite{pu2016variational} or GANs~\cite{goodfellow2014generative}. Thus, we leverage the phase information to detect up-sampling artifacts. 
% For applying phase information to CNNs, we reconstruct the spatial domain representation of the phase spectrum from the frequency domain by Inverse Discrete Fourier Transform~(IDFT) with the frequency spectrum without amplitude. 
For applying phase information to CNNs, we reconstruct the spatial domain representation of the phase spectrum from the frequency domain (\ie IDFT with the frequency spectrum without amplitude).
Finally, we concatenate the spatial domain representation of the phase spectrum with the RGB image in the channel, which results in an RGBP 4-channel image.

\subsection{Capturing up-sampling artifacts via phase spectrum in face forgery}\label{sec:derivation_3}

To detect the observed common artifacts, namely up-sampling, we analyze it in the frequency domain.

% Up-sampling is a essential step of forged face generation with both AutoEncoders and GANs, and it 
Up-sampling will lead to the emergence of new frequency components. And we make a claim as follows
\begin{claim}
% The introduction of phase spectrum makes up the disadvantage of amplitude, and it remarkably helps face forgery detection.
% Phase spectrum is more easily capturing up-sampling artifacts and therefore helps face forgery detection.
Phase spectrum is more sensitive to up-sampling artifacts and therefore helps face forgery detection.
\end{claim}
To simplify the calculation, we make all the mathematical derivation based on the 1D signal. We first set up the basic notations used in this paper: $x(n)$ and $\mathbf{X}(u)$ denote a 1D discrete signal and its Discrete Fourier Transform(DFT), where $n$ is the spatial location of the signal and $u$ represents the frequency. $\mathbf{A}(u)$ is the amplitude spectrum and $\mathbf{P}(u)$ is the phase spectrum. And we use $c(n)$ and $\mathbf{C}(n)$ to denote the convolution kernel and its DFT representation. And we use $\ast$ to denote convolutional operation. 
% Besides, $\mathbf{F}(\cdot)$ and $\mathbf{F}^{-1}(\cdot)$ represent the DFT and its inverse. So we have $\mathbf{X}(u)=\mathbf{F}(x(n))$ and $x(n)=\mathbf{F}^{-1}(\mathbf{X}(u))$.
\begin{proof}
The increase of spatial resolution in 2D corresponds to the extension of the time domain in 1D. Assume that the input $x(n)$ is up-sampled by factor 2, then
\begin{equation}
\hat{x}(n)=
\begin{cases}
x(\frac{1}{2}n), & n = 2k\\
0, & n=2k+1
\end{cases}
\end{equation}
where $k=0,1,2,\cdots,N-1$, and
\begin{equation} \label{eq:freq_comp}
\begin{aligned}
\hat{\mathbf{X}}(u)&=\frac{1}{2N}\sum_{n=0}^{2N-1}\hat{x}(n)e^{-j\frac{2\pi un}{2N}} \\
                    &=\frac{1}{2N}\sum_{n=0}^{N-1}\hat{x}(2n)e^{-j\frac{2\pi u2n}{2N}} \\
                    &=\frac{1}{2N}\sum_{n=0}^{N-1}x(n)e^{-j\frac{2\pi 2un}{N}} \\
                    % &=\underbrace{a_0+a_{1}e^{j\theta_{1}}+\cdots+a_{2N-1}e^{j\theta_{2N-1}}}_{2N\ \text{items}}
                    % &=a_0+a_{1}e^{j\theta_{1}}+\cdots+a_{2N-1}e^{j\theta_{2N-1}}
\end{aligned}
\end{equation}

The we have $\hat{x}(n)=x(\frac{1}{2}n)\Leftrightarrow\hat{\mathbf{X}}(u)=\mathbf{X}(2u)$ with the Eq.~\ref{eq:freq_comp}, which leads to the conclusion that the increase of spatial resolution will result in the compression in the frequency domain which is consistent with the property of Fourier Transform (FT). In fact, the essence of DFT is the principle value interval of Discrete Fourier Series (DFS) and thus new frequency components are the duplicate of origin frequency components.

Base on our inference that phase spectrum will keep more frequency components that tend to zero in amplitude spectrum, which is detailedly proved in the \textbf{Appendix}~\textcolor{red}{1.1}. We first assume the amplitude spectrum $\mathbf{X_A}(u)$ and the phase spectrum $\mathbf{X_P}(u)$ of original images $x(n)$. It is
\begin{equation}
\begin{aligned}
% \mathbf{X}_A(u) &= a_0+a_1e^{j\theta_1}+\cdots+a_{N-1}e^{j\theta_{N-1}} \\
\mathbf{X}_A(u) &= \underbrace{a_0+a_1e^{j\theta_1}+\cdots+a_{k}e^{j\theta_{k}}}_{(k+1)\ \text{items}} \\
\mathbf{X}_P(u) &= \underbrace{p_0+p_1e^{j\theta_1}+\cdots+p_{N-1}e^{j\theta_{N-1}}}_{N\ \text{items}} \\
\end{aligned}
\end{equation}
and the corresponding up-sampling is
\begin{equation}
\begin{aligned}
% \mathbf{X}_A^{up}(u) &= a_0+a_1e^{j\theta_1}+\cdots+a_{2N-1}e^{j\theta_{2N-1}} \\
\mathbf{X}_A^{up}(u)\!&=\!\underbrace{a_0\!+\!\cdots\!+\!a_{k}e^{j\theta_{k}}}_{(k+1)\ \text{items}}\!+\!\underbrace{a_Ne^{j\theta_N}\!+\!\cdots\!+\!a_{N\!+\!k}e^{j\theta_{N\!+\!k}}}_{(k+1)\ \text{items}} \\
\mathbf{X}_P^{up}(u) &= \underbrace{p_0+p_1e^{j\theta_1}+\cdots+p_{2N-1}e^{j\theta_{2N-1}}}_{2N\ \text{items}}
\end{aligned}
\end{equation}

We define that $y_A(n)$ is the output of a convolution layer with an input $x(n)$ and its frequency domain form is $\mathbf{Y_A}(u)$. And we get
\begin{equation}
\begin{aligned}
y_A(n) =&\ x(n)\ast c(n) \\
&\Updownarrow \\
\mathbf{Y}_A(u) = &\ \mathbf{X_A}(u) \cdot \mathbf{C}(u)
\end{aligned}
\end{equation} 
According to the deduction that the phase spectrum helps CNNs acquire and learn more abundant frequency components which are ignored with convolution calculations of amplitude spectrum proved in \textbf{Appendix} \textcolor{red}{1.2}, we can deduce the frequency domain form is
\begin{equation}
\mathbf{Y}_A(u) = \underbrace{f_0+f_1e^{j\theta_1}+\cdots+f_{k+N-1}e^{j\theta_{k+N-1}}}_{k+N\ \text{items}}
\end{equation}
% And the corresponding form of the up-sampling is
\begin{equation}
\mathbf{Y}_A^{up}(u) = \underbrace{f_0+f_1e^{j\theta_1}+\cdots+f_{2N+k-1}e^{j\theta_{2N+k-1}}}_{2N\ \text{items}}
\end{equation}

In our work, we first take Inverse Discrete Fourier Transform~(IDFT) to phase spectrum and acquire the spatial domain form $p(n)$ of phase. And we state a theorem named \textit{the distributive law} as follow,
\begin{theorem}\label{theo:distributive law}
$(f(\cdot)+g(\cdot))\ast h(\cdot) = f(\cdot)\ast h(\cdot) + g(\cdot)\ast h(\cdot)$
\end{theorem}

Then we consider that we directly concatenate $x(n)$ and $p(n)$ in channel dimension based on theorem~\ref{theo:distributive law} and the output $\mathbf{Y}_{A+P}(u)$ is
\begin{equation}
\mathbf{Y}_{A+P}(u) = \underbrace{f_0+f_1e^{j\theta_1}+\cdots+f_{2N-2}e^{j\theta_{2N-2}}}_{2N-1\ \text{items}}
\end{equation}
% And the corresponding form of the up-sampling is
\begin{equation}
\mathbf{Y}_{A+P}^{up}(u) = \underbrace{f_0+f_1e^{j\theta_1}+\cdots+f_{3N-2}e^{j\theta_{3N-2}}}_{3N-1\ \text{items}}
\end{equation}
\end{proof}

Intuitively, the number of learnable frequency components is $N$ when we leverage the original image and its phase together, but the number just is $N-k$ with the utilization of the original image alone. Therefore, it is clear that the difference between $\mathbf{Y}_{A+P}(u)$ and $\mathbf{Y}_{A+P}^{up}(u)$ is bigger than $\mathbf{Y}_{A}(u)$ and $\mathbf{Y}_{A}^{up}(u)$. In particular, the value of $k$ is usually small in nature images and thus our method observably improves the performance on the detection of face forgery. 
Thus, we conclude that the introduction of the phase spectrum helps capture more frequency artifacts caused by cumulative up-sampling in deepfake video generation.

\subsection{Suppressing the semantic information and focusing on local region}~\label{sec:shallow}
We consider that the pivotal distinction between pristine face and forged face is local low-level features(\eg textures, colors) instead of global high-level semantic features(\eg face, human). Because most of these semantic features are shared in both pristine and forged faces, extensive high-level semantic information more or less has a negative effect on forged face detection as it contains many common characteristics of pristine and forged face images. 
% General CNNs is normally very deep and result in large receptive fields which make better access to high-level features. 
For the purpose of suppressing high-level semantic features and extracting more texture features, we straightforwardly shallow the neural network by throwing away many convolutional layers or blocks. Then we demonstrate that shallow networks are more transferable and efficient simultaneously.

\begin{table}[tb]
\setlength{\belowcaptionskip}{-0.3cm}
\centering
\begin{tabular}{c|cc|cc}
\toprule
% Methods& ACC& AUC& ACC& AUC \\
    %   &(HQ)&(HQ)&(LQ)&(LQ)\\
\multirow{2}{*}{Methods} & \multicolumn{2}{c|}{HQ} & \multicolumn{2}{c}{LQ} \\ \cline{2-5}
    & ACC        & AUC       & ACC        & AUC    \\ 
\midrule
Steg. Features~\cite{fridrich2012rich}& 70.97& -& 55.98& -\\
Cozzolino~\etal ~\cite{cozzolino2017recasting}& 78.45& -& 58.69& -\\
Bayer \& Stamm~\cite{bayar2016deep}& 82.97& -& 66.84& -\\
Rahmouni~\etal ~\cite{rahmouni2017distinguishing}& 79.08& -& 61.18& -\\
MesoNet~\cite{afchar2018mesonet}& 83.10& -& 70.47& -\\
Face X-ray~\cite{li2020face}& -& 87.35& -& 61.60\\
Xception~\cite{Chollet_2017_CVPR}& \textbf{92.39}& 94.86& 80.32& 81.76\\
\midrule
Ours(Xception)& 91.50& \textbf{95.32}& \textbf{81.57}& \textbf{82.82} \\
\bottomrule
\end{tabular}
\caption{Quantitative results~(ACC~(\%) and AUC~(\%)) on FaceForensics++ dataset with high-quality~(light compression) and low quality~(heavy compression) settings. The bold results are the best.}
\label{tab:c23_c40}
\end{table}

\section{Experiments}
In this section, we first introduce the overall experimental settings and then present extensive experimental results to demonstrate the superiority of our approach.
\subsection{Experimental settings}
\textbf{Datasets.}
Following recent related works~\cite{10.1007/978-3-030-58610-2_6,masi2020two,li2020face,li2020sharp} of face forgery detection, we conduct our experiments on the two benchmark public deepfake datasets: FaceForensics++(FF++)~\cite{rossler2019faceforensics++} and Celeb-DF~\cite{li2020celeb}. Both of them are large-scale and contain pristine and manipulated videos of human faces. FF++ consists of four kinds of common face manipulation methods~\cite{DeepFakes,thies2016face2face,FaceSwap,thies2019deferred}. Celeb-DF is in general the most challenging to the current detection methods, and their overall performance on Celeb-DF is lowest across all datasets.

\textbf{Evaluation metrics.} 
In our experiments, we mainly utilize the Accuracy rate (ACC) and the Area Under Receiver Operating Characteristic Curve (AUC) as our evaluation metrics. \textbf{(1) ACC.} Accuracy rate is the most intuitive metric in face forgery detection. It is also applied to FF++~\cite{rossler2019faceforensics++} and thus we use ACC as the major evaluation metric in the experiment. \textbf{(2) AUC.} Following the Celeb-DF~\cite{li2020celeb} and Two-branch~\cite{masi2020two}, we use AUC as another evaluation metric to evaluate the performance on cross-dataset. Besides, we use the recall rate as our multi-class classification evaluation metric.

\textbf{Implementation and Hyper-Parameters.}
In our experiments, we use Xception~\cite{Chollet_2017_CVPR} as the backbone of our approach. For the purpose of reducing receptive fields, we just retain the Xception Block 1-3
% Xception Block 3 
and Xception Block 12. The final spatial form of our phase spectrum is the IDFT of the absolute value of the pristine phase spectrum. We optimize the networks by Adam optimizer~\cite{kingma2014adam}. The initial learning rate $lr=2\times10^{-3}$ and it drops to half of itself every time the validation loss does not decrease after 5 full epochs.

\begin{figure*}[t]
\centering
\includegraphics[scale=0.57]{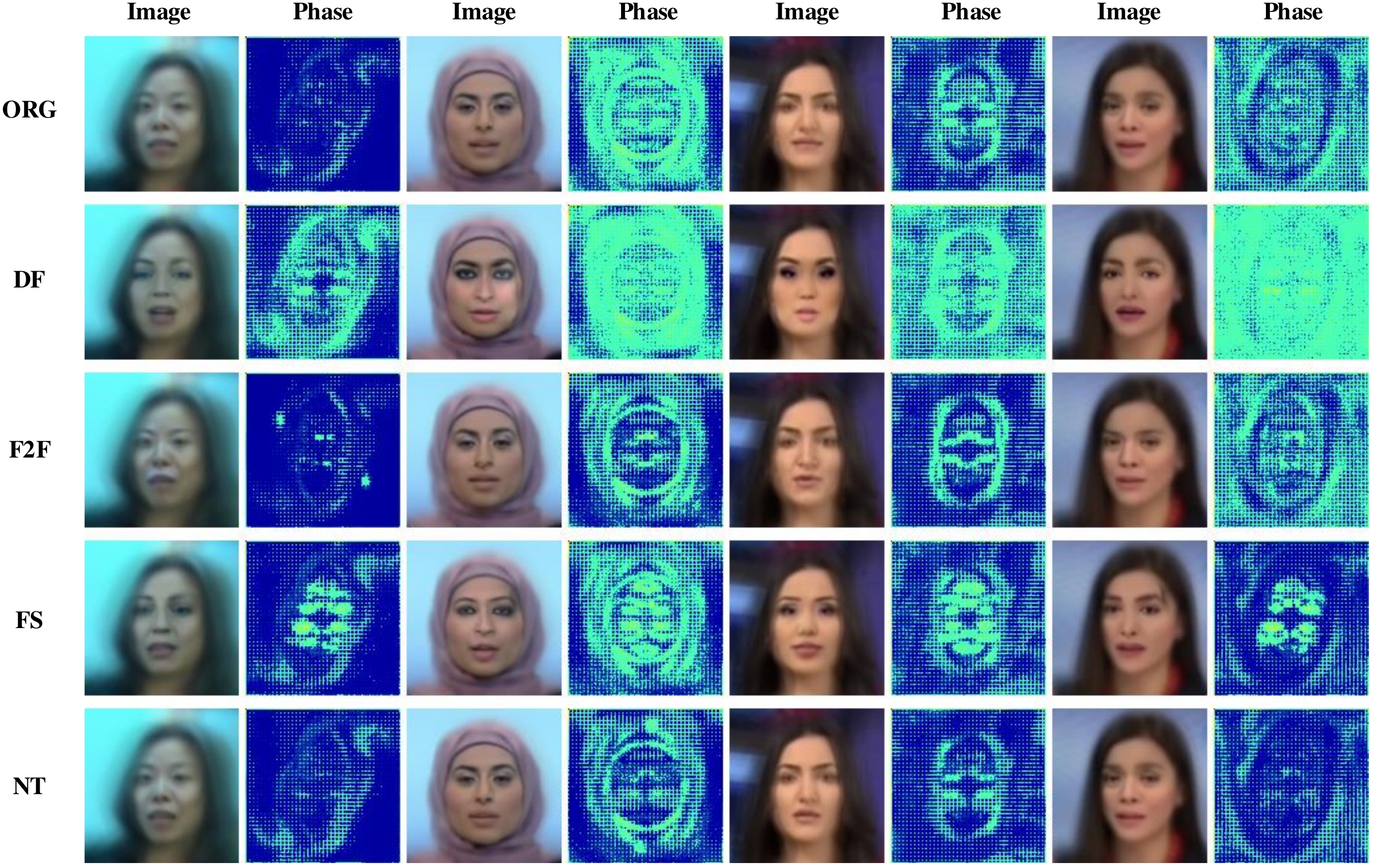}
\caption{Visualization of various manipulation methods and our spatial domain representations of the phase spectrum in FF++~\cite{rossler2019faceforensics++}. Each image and phase spectrum is the average of all frames of a video. Every manipulation method tends to be a specific pattern in the phase spectrum while it is not obvious in the RGB domain. Best viewed in color.}
\label{fig:aver-phase}
\end{figure*}

\subsection{Comparison with previous methods}
In this section, we compare our method with previous deepfake detection methods. We train all models on only FF++~\cite{rossler2019faceforensics++} and respectively evaluate them on FF++ in Section~\ref{exp:ff++} and Celeb-DF in Section~\ref{exp:celeb}.
\subsubsection{Comparable results on FF++}\label{exp:ff++}
In this section, we compare our method with previous deepfake detection methods on FF++~\cite{rossler2019faceforensics++}. Although our primary purpose is to improve the generalization and transferability, we also obtain comparable results in FF++~\cite{rossler2019faceforensics++}. We first evaluate our methods on different video compression settings including high quality~(HQ~(c23)) and low quality~(LQ~(c40)). As the results shown in Table~\ref{tab:c23_c40}, the proposed method outperforms or is on par with baseline in both ACC and AUC with LQ settings. Low-quality videos have been highly compressed and many frequency components are weakened. The improvement of performance mainly benefits from the extra phase information captured by CNNs, which keeps more frequency components than plain RGB-based images. At the same time, we also obtain comparable results 
% compared to other methods 
with HQ settings.

Furthermore, we also evaluate our approach on different face manipulation methods in FF++~\cite{rossler2019faceforensics++}. The results are demonstrated in Table~\ref{tab:ffpp}. We train and test our models exactly on low-quality videos for each manipulation methods. We also reproduced the results of MesoNet~\cite{afchar2018mesonet} and Xception~\cite{Chollet_2017_CVPR}, and other results are directly cited from~\cite{rossler2019faceforensics++}. In general, basic experiments also show comparable results with previous methods though the transferability is the main purpose of SPSL.

\begin{table*}[t]
\setlength{\belowcaptionskip}{-0.2cm}
\centering
\setlength{\tabcolsep}{3.8mm}{
\begin{tabular}{c|cc|cc|cc|cc}
\toprule
\multirow{2}{*}{Methods} & \multicolumn{2}{c|}{DF~\cite{DeepFakes}} & \multicolumn{2}{c|}{F2F~\cite{thies2016face2face}} & \multicolumn{2}{c|}{FS~\cite{FaceSwap}} & \multicolumn{2}{c}{NT~\cite{thies2019deferred}}  \\ \cline{2-9}
 & ACC & AUC & ACC & AUC & ACC & AUC & ACC & AUC\\
\midrule
Steg. Features~\cite{fridrich2012rich}& 73.64 & -& 73.72 & -& 68.93 & -& 63.33 & -\\
Cozzolino \etal ~\cite{cozzolino2017recasting}& 85.45 & -& 67.88 & -& 73.79 & -& 78.00 & -\\
Rahmouni \etal ~\cite{rahmouni2017distinguishing}& 85.45 & -& 64.23 & -& 56.31 & -& 60.07 & -\\
Bayar and Stamm~\cite{bayar2016deep}& 84.55 & -& 73.72 & -& 82.52 & -& 70.67 & -\\
MesoNet~\cite{afchar2018mesonet}& 87.27 & -& 56.20 & -& 61.17 & -& 40.67 & -\\
XceptionNet~\cite{Chollet_2017_CVPR}& \textbf{95.15} & \textbf{99.08}& 83.48 & 93.77& 92.09 & 97.42& \textbf{77.89} & \textbf{84.23}\\
\midrule
Ours(Xception)& 93.48 & 98.50& \textbf{86.02} & \textbf{94.62}& \textbf{92.26} & \textbf{98.10}& 76.78 & 80.49\\
\bottomrule 
\end{tabular}}
\caption{Quantitative results~(ACC~(\%) and AUC~(\%)) on FaceForensics++ dataset with four different manipulation methods,~\ie DeepFakes(DF)~\cite{DeepFakes}, Face2Face(F2F)~\cite{thies2016face2face}, FaceSwap(FS)~\cite{FaceSwap}, NeuralTextures(NT)~\cite{thies2019deferred}. The bold results are best.}
\label{tab:ffpp}
\end{table*}

\subsubsection{Cross-dataset evaluation on Celeb-DF}\label{exp:celeb}
In this section, we evaluate the transferability of our method given that it is trained on FF++ with multiple manipulations but tested on Celeb-DF. We first verify that most of the previous methods show a drastic performance drop on the cross-dataset evaluation. Table~\ref{tab:transfer} shows the AUC comparison with some recent methods for face forgery detection. Our method obtains the state-of-the-art AUC on Celeb-DF while still having a good performance on FF++ compared with all the other methods. The performance gains mainly benefit from the extra appreciable frequency components of the phase spectrum, which are enhanced many times in the cumulative up-sampling, and thus the differences of frequency components perceptible by convolutional kernel between pristine images and forgery become more striking. Therefore, the proposed SPSL is capable of detecting common artifacts.
% Thus, convolutional layers can more easily capture the changes in the frequency domain between original and fake faces with the introduction of the phase spectrum. At the same time, almost all forgery is generated by AutoEncoder or GANs and therefore phase spectrum takes advantage of the common characteristic to obtain transferability.
% Moreover, a shallow network makes our method more robust by suppressing the disadvantageous common high-level information in different datasets. 

\begin{table}[htb]
\centering
\begin{tabular}{ccc}
\toprule
Method& FF++~\cite{rossler2019faceforensics++}& Celeb-DF~\cite{li2020celeb} \\
\midrule
Two-stream~\cite{zhou2017two}& 70.10& 53.80\\
Meso4~\cite{afchar2018mesonet}& 84.70& 54.80\\
MesoInception4~\cite{afchar2018mesonet}& 83.00& 53.60\\
HeadPose~\cite{yang2019exposing}& 47.30& 54.60\\
FWA~\cite{li2018exposing}& 80.10& 56.90\\
VA-MLP~\cite{matern2019exploiting}& 66.40& 55.00\\
VA-LogReg& 78.00& 55.10\\
% Xception-raw~\cite{rossler2019faceforensics++}& \textbf{99.7}& 48.2\\
% Xception-c23~\cite{rossler2019faceforensics++}& \textbf{99.7}& 65.3\\
Xception-c40~\cite{rossler2019faceforensics++}& 95.50& 65.50\\
Multi-task~\cite{nguyen2019multi}& 76.30& 54.30\\
Capsule~\cite{nguyen2019capsule}& 96.60& 57.50\\
DSP-FWA~\cite{li2018exposing}& 93.00& 64.60\\
SMIL~\cite{li2020sharp}&     96.80& 56.30\\
Two-branch~\cite{masi2020two}& 93.20& 73.40\\
$\text{F}^3\text{-Net}$~\cite{10.1007/978-3-030-58610-2_6}&  \textbf{97.97}& 65.17\\
\midrule
% \midrule
% Xception-Phase& 87.89& 67.30\\
% % \midrule
% Xception-Shallow& 81.80& 67.55\\
% \midrule
SPSL(Xception)& 96.91& \textbf{76.88}\\
\bottomrule
\end{tabular}
\caption{\textbf{Cross-dataset evaluation (AUC (\%) ) on Celeb-DF}. Best competing methods on Celeb-DF are reported. Our method obtains the state-of-the-art performance on cross-dataset evaluation. At the same time, our method still performs well when tested on just deepfake class(96.91\%) AUC on FF++. Results for some other methods are from~\cite{li2020celeb}, and the bold results are the best.}
\label{tab:transfer}
\end{table}

\subsection{Multi-class classification evaluation}
\begin{figure}[t]
\setlength{\belowcaptionskip}{-0.3cm}
% \centering
\subfigure[Baseline]{
	\begin{minipage}{3.95cm}
	\centering
	\includegraphics[scale=0.35]{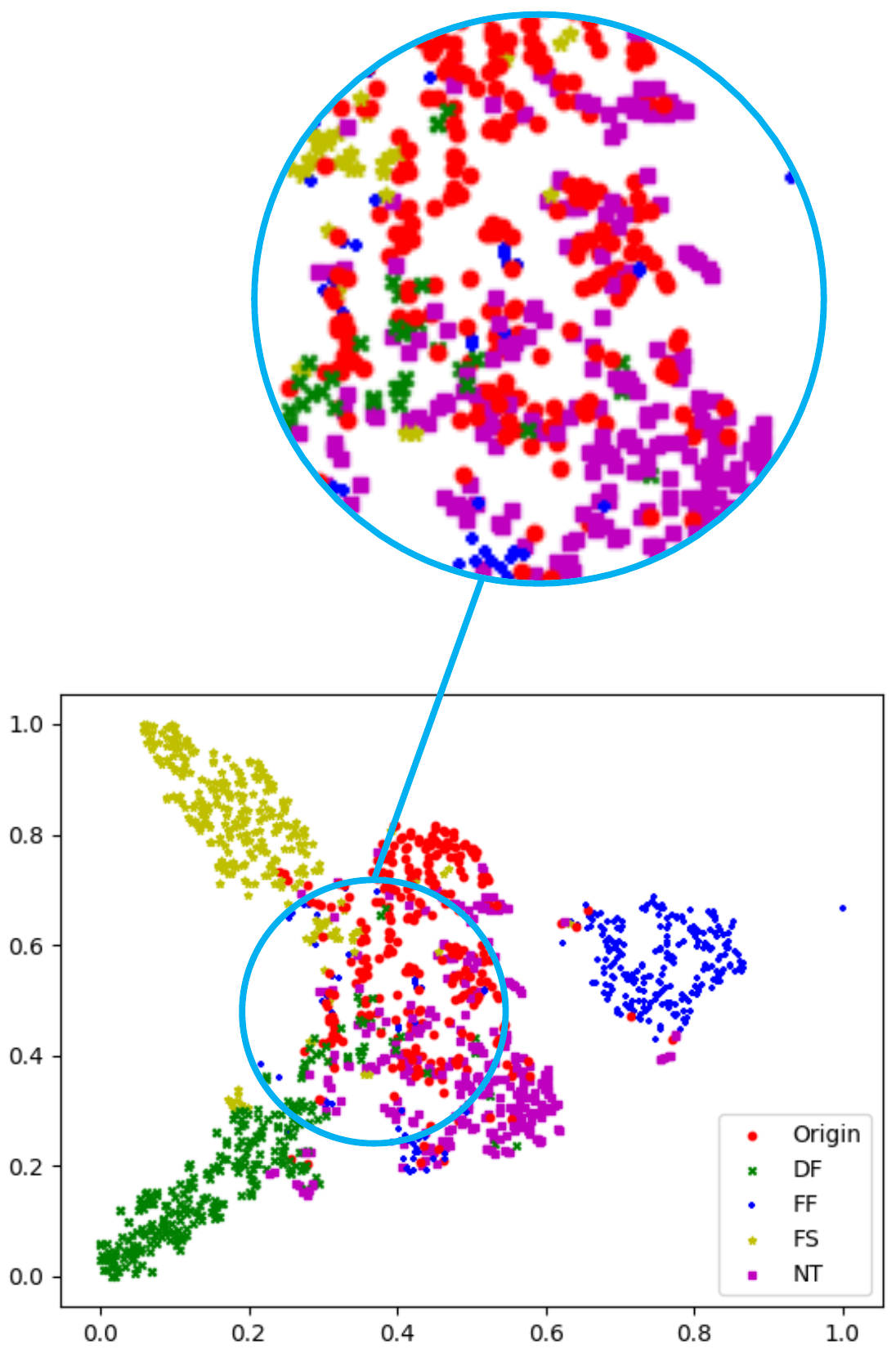}
	\end{minipage}
	\label{fig:t-SNE-baseline}
}
\subfigure[SPSL]{
	\begin{minipage}{3.95cm}
	\centering
	\includegraphics[scale=0.35]{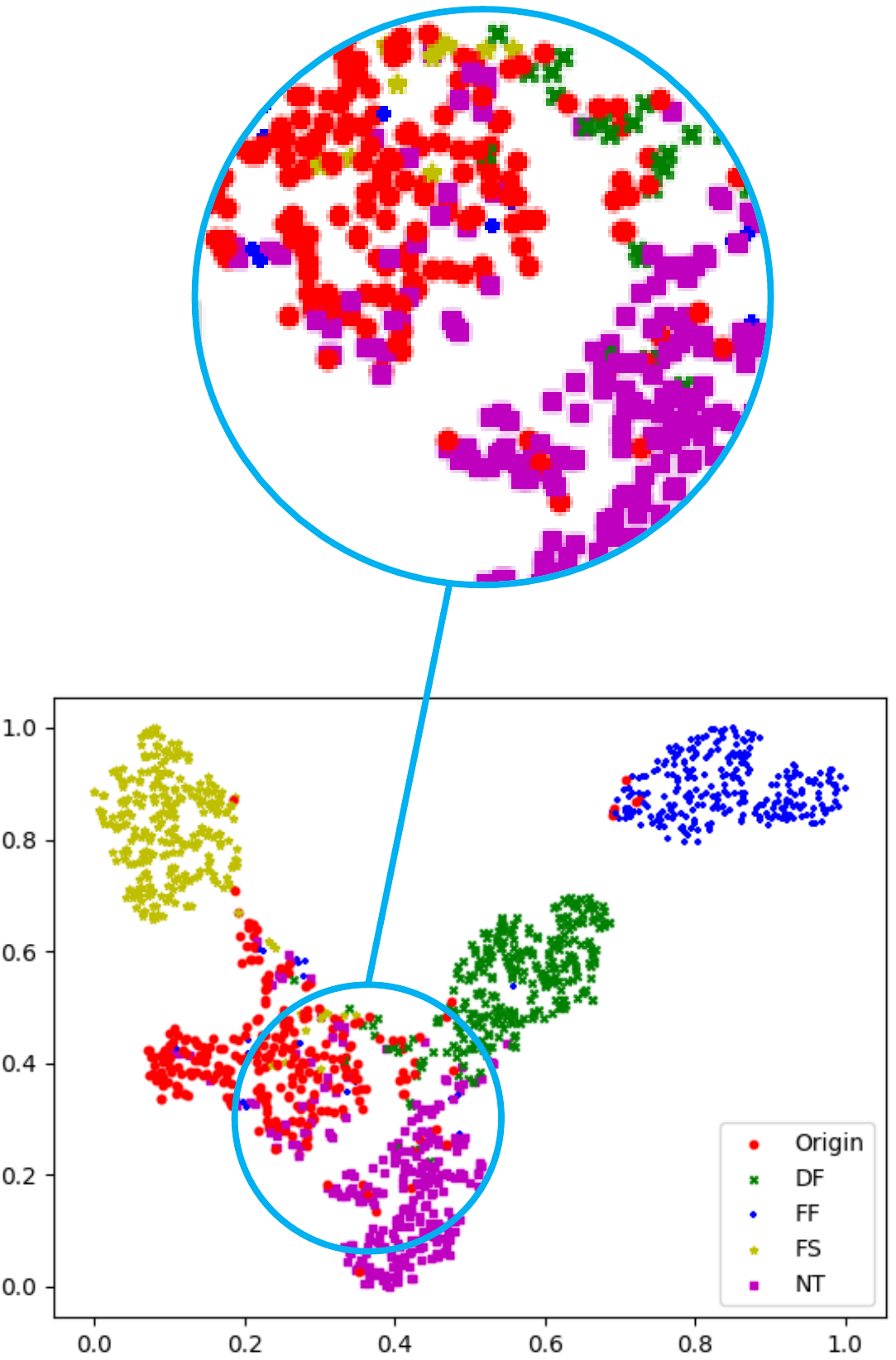}
	\end{minipage}
	\label{fig:t-SNE-ours}
}
\caption{The t-SNE feature spaces visualization of the basic Xception~(a) and SPSL~(b) on FaceForensics++~\cite{rossler2019faceforensics++} high quality (HQ) in the multi-class classification task. Red color dots represent pristine images, and rest colors respectively indicate the different manipulation methods. Best viewed in color.}
\label{fig:t-SNE}
\end{figure}

We furthermore evaluate the proposed SPSL on multi-class classification with different face manipulation methods list in FF++~\cite{rossler2019faceforensics++} in this section. The models are trained and tested on FF++ with five types of labels, and multi-class classification is more challenging and significant than binary classification. The results are shown in Table~\ref{tab:mul-class} by the way of recall rate. With all three kinds of compression setting, the proposed SPSL completely surpasses the original XceptionNet method. Furthermore, we also show the t-SNE~\cite{maaten2008visualizing} feature spaces of data in FF++ high quality with the multi-class classification task, by the Xception and our SPSL, as shown in Figure~\ref{fig:t-SNE}. Xception is more likely to confuse pristine faces with NeuralTextures-based fake faces because this manipulation method modifies very limited pixels in the spatial domain, as shown in Figure~\ref{fig:t-SNE-baseline}. Conversely, the proposed SPSL can split up all classes in the embedding feature spaces, as shown in Figure~\ref{fig:t-SNE-ours}. These improvements may benefit from the salient difference of phase spectrum among various manipulation methods and we show the average phase spectrum in the spatial domain of every frame of a video in Figure~\ref{fig:aver-phase}. For all four kinds of manipulation methods in FF++~\cite{rossler2019faceforensics++}, the spatial images of the phase spectrum show distinguishing results for each method. In particular, NeuralTextures-based images, which just slightly tampered with lip, are very similar to pristine images causing almost indistinguishable in the RGB domain but the spatial images of the phase spectrum are still separable.

\begin{table}[htb]
\centering
\setlength{\tabcolsep}{1.5mm}{
\begin{tabular}{c|c|c|c|c|c}
\toprule
Methods& DF& F2F& FS& NT& ORG \\
\midrule
MesoIncep4~\cite{afchar2018mesonet}& 94.81&  43.32& 73.24& 40.39& 85.16\\
Xception-c0~\cite{Chollet_2017_CVPR}& 97.84& 96.68& 96.84& 87.67& 98.03\\
\hline
SPSL& \multirow{2}*{\textbf{99.05}}&  \multirow{2}*{\textbf{97.20}}&  \multirow{2}*{\textbf{97.63}}&  \multirow{2}*{\textbf{91.40}}&  \multirow{2}*{\textbf{98.25}}\\
(Xception-c0)& & & & & \\
\midrule
\midrule
Xception-c23~\cite{Chollet_2017_CVPR}& 88.00& 88.61& 87.07& 74.83& 75.52\\
\hline
SPSL&  \multirow{2}*{\textbf{94.18}}&  \multirow{2}*{\textbf{93.59}}&  \multirow{2}*{\textbf{95.62}}&  \multirow{2}*{\textbf{81.72}}&  \multirow{2}*{\textbf{88.72}}\\
(Xception-c23)& & & & & \\
\midrule
\midrule
Xception-c40~\cite{Chollet_2017_CVPR}& 86.61&  \textbf{78.88}& 83.16& 52.94& 75.55\\
\hline
SPSL&  \multirow{2}*{\textbf{91.16}}& \multirow{2}*{78.31}&  \multirow{2}*{\textbf{88.75}}&  \multirow{2}*{\textbf{58.97}}&  \multirow{2}*{\textbf{77.49}}\\
(Xception-c40)& & & & & \\
\bottomrule
\end{tabular}}
\caption{The recall rate (\%) of origin and each manipulation method with Raw~(c0) , HQ~(c23) and LQ~(c40) settings in our multi-class classification.}
\label{tab:mul-class}
\end{table}

\begin{figure*}[t]
\setlength{\belowcaptionskip}{-0.2cm}
\centering
\includegraphics[scale=0.47]{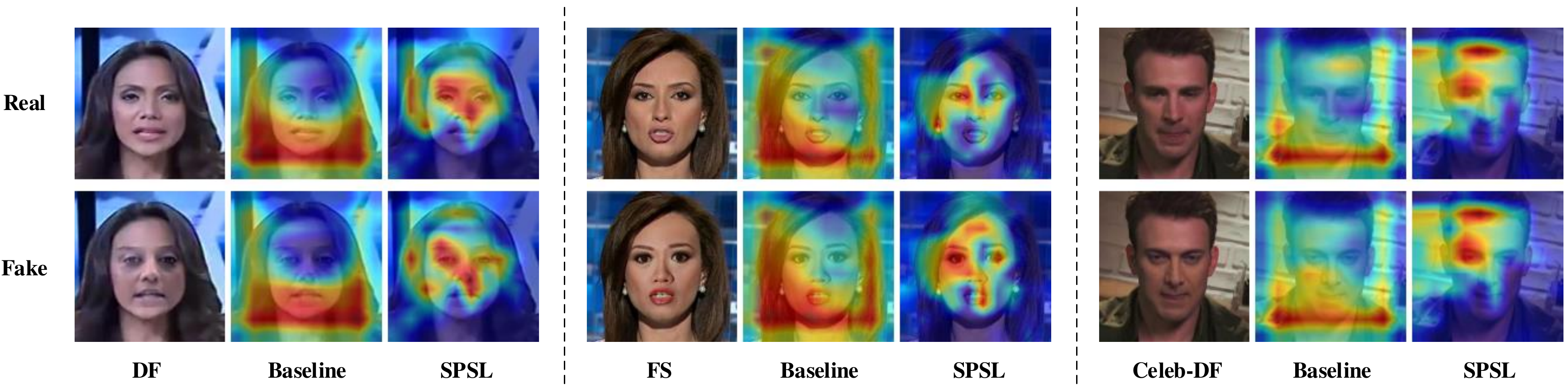}
\caption{The Grad-CAM of the baseline Xception and the proposed SPSL, including two different manipulation methods in FF++~\cite{rossler2019faceforensics++} and another Celeb-DF~\cite{li2020celeb} datasets. Best viewed in color.}
\label{fig:gradcam}
\end{figure*}

\section{Ablation Study}
\subsection{Effectiveness of Phase spectrum and Shallow network}
To evaluate the effectiveness of both Phase spectrum and Shallow network, we first respectively evaluate one of them with baseline and combine them finally. All models are trained on FF++~\cite{rossler2019faceforensics++} and tested on Celeb-DF~\cite{li2020celeb}. The results are listed in Table~\ref{tab:ablation}. Compared with model 1(baseline Xception), model 2(Xception with phase spectrum) and model 3 (shallow Xception) improve the AUC scores of Celeb-DF. The transferability has a great improvement with both of them. When combining phase spectrum and shallow network, SPSL gets the best performance and the AUC score increase by about 13\%. Furthermore, to demonstrate the effectiveness of the proposed SPSL better, we respectively visualize the Gradient-weighted Class Activation Mapping(Grad-CAM)~\cite{selvaraju2017grad} of the baseline and SPSL, as shown in Figure~\ref{fig:gradcam}. The Grad-CAM indicates that the proposed SPSL prefers to focus on more microcosmic regions while the baseline model pays more attention to global information, and this phenomenon also accords with our motivation. Furthermore, we demonstrate the correlation analysis between the performance and the number of convolution layers of various backbone networks in \textbf{Appendix} \textcolor{red}{2}.

\begin{table}[htb]
\setlength{\belowcaptionskip}{-0.3cm}
\centering
\setlength{\tabcolsep}{3mm}{
\begin{tabular}{c|cc|c}
\toprule
ID& Phase& Shallow& Celeb-DF~\cite{li2020celeb}\\
\hline
1&-   &-      & 59.98\\
2&$\surd$ &-& 69.01\\
3&-   &$\surd$& 66.74\\
4&$\surd$&$\surd$& \textbf{72.39}\\
\bottomrule
\end{tabular}}
\caption{Ablation study of the proposed SPSL. These models are trained on FF++ with high quality(HQ) settings and tested on Celeb-DF (AUC (\%) ). We compare SPSL and its variants by removing phase spectrum and shallow operation step by step.}
\label{tab:ablation}
\end{table}

\subsection{Universality of SPSL with Various Backbones}
All the above-mentioned experiments are based on XceptionNet~\cite{Chollet_2017_CVPR}, and thus we also evaluate the universality of SPSL with two types of ResNet~\cite{he2016deep}. For both ResNet34 and ResNet50, we directly halve the residual block to shallow networks. The results listed in Table~\ref{tab:backbone} demonstrate that the proposed SPSL is a general framework for various backbones.

\begin{table}[htb]
\setlength{\belowcaptionskip}{-0.3cm}
\centering
\setlength{\tabcolsep}{2mm}{
\begin{tabular}{c|c|c|c|c}
\toprule
\multirow{2}{*}{Backbone}&  \multicolumn{2}{c|}{FF++~\cite{rossler2019faceforensics++}} & \multicolumn{2}{c}{Celeb-DF~\cite{li2020celeb}} \\ \cline{2-5}
        & ACC& AUC& ACC& AUC\\
\midrule
\midrule
ResNet-34& 71.55& 81.58& 65.19& 66.90\\
\midrule
SPSL& \multirow{2}*{\textbf{83.24}}& \multirow{2}*{\textbf{89.26}}& \multirow{2}*{\textbf{66.79}}& \multirow{2}*{\textbf{71.78}} \\
(ResNet-34)& & & & \\
\midrule
\midrule
ResNet-50& 81.83& 83.51& \textbf{69.40}& 70.05\\
\midrule
SPSL& \multirow{2}*{\textbf{86.64}}& \multirow{2}*{\textbf{91.04}}& \multirow{2}*{68.28}& \multirow{2}*{\textbf{73.09}} \\
(ResNet-50)& & & & \\
\bottomrule
\end{tabular}}
\caption{The results (ACC (\%) and AUC (\%) ) on FF++~\cite{rossler2019faceforensics++} and cross-dataset evaluation on Celeb-DF~\cite{li2020celeb} of two different backbones with the proposed SPSL.}
\label{tab:backbone}
\end{table}

\section{Limitations}
Even if we have demonstrated the effectiveness of the proposed SPSL and achieved satisfactory performance on cross-dataset evaluation and multi-class classification, we are aware that there exist some limitations of our work.

Our method depends on the existence of up-sampling in forged face generation. Thus, the performance may drop if the forgery face is not produced by methods based on generative models. Besides, our method also suffers from a transferability drop when encountering an entirely different type of face forgery manipulation. For instance, the model trained on identity swap datasets may fail to detect forgery faces whose expressions are swapped. This is expected since manipulations from different categories can leave a specific trace in the phase spectrum as shown in Figure~\ref{fig:aver-phase}, this is also the reason why our method makes a remarkable improvement of multi-class classification.

\section{Conclusion}
In this work, we propose a novel face forgery detection method, SPSL, which takes advantage of both spatial and frequency information. The core competence of SPSL is that phase spectrum contains more abundant appreciable frequency components and these components will be duplicated in the process of up-sampling which is the necessary step of forged face generation. Besides, SPSL forces the network to focus on the local microcosmic region and suppress global semantic information for more robustness. We perform a meticulous mathematical derivation to prove the rationality of the proposed SPSL, and extensive experiments demonstrate that the SPSL has an excellent performance on the face forgery detection, especially in the challenging cross-dataset evaluation task.

% \section{Acknowledgement}
% \textbf{Acknowledgement.} 
% \noindent \textbf{Acknowledgement} This work was supported in part by the Natural Science Foundation of China under Grant U20B2047, U1636201, 62002334, by the Anhui Science Foundation of China under Grant 2008085QF296, by the Exploration Fund Project of the University of Science and Technology of China under Grant YD3480002001 and the Fundamental Research Funds for the Central Universities under Grant WK2100000011.

\newpage

{\small
\bibliographystyle{ieee_fullname}
\bibliography{HongguLiu_Spatial_Phase_Shallow_Learning__Rethinking_Face_Forgery_Detection_in_Frequency_Domain}

\begin{thebibliography}{10}\itemsep=-1pt

\bibitem{afchar2018mesonet}
Darius Afchar, Vincent Nozick, Junichi Yamagishi, and Isao Echizen.
\newblock Mesonet: a compact facial video forgery detection network.
\newblock In {\em 2018 IEEE International Workshop on Information Forensics and
  Security (WIFS)}, pages 1--7. IEEE, 2018.

\bibitem{agarwal2019protecting}
Shruti Agarwal, Hany Farid, Yuming Gu, Mingming He, Koki Nagano, and Hao Li.
\newblock Protecting world leaders against deep fakes.
\newblock In {\em CVPR Workshops}, pages 38--45, 2019.

\bibitem{araujo2019computing}
André Araujo, Wade Norris, and Jack Sim.
\newblock Computing receptive fields of convolutional neural networks.
\newblock {\em Distill}, 2019.
\newblock https://distill.pub/2019/computing-receptive-fields.

\bibitem{bayar2016deep}
Belhassen Bayar and Matthew~C Stamm.
\newblock A deep learning approach to universal image manipulation detection
  using a new convolutional layer.
\newblock In {\em Proceedings of the 4th ACM Workshop on Information Hiding and
  Multimedia Security}, pages 5--10, 2016.

\bibitem{chai2020makes}
Lucy Chai, David Bau, Ser-Nam Lim, and Phillip Isola.
\newblock What makes fake images detectable? understanding properties that
  generalize.
\newblock {\em arXiv preprint arXiv:2008.10588}, 2020.

\bibitem{Chollet_2017_CVPR}
Francois Chollet.
\newblock Xception: Deep learning with depthwise separable convolutions.
\newblock In {\em Proceedings of the IEEE Conference on Computer Vision and
  Pattern Recognition (CVPR)}, July 2017.

\bibitem{cozzolino2017recasting}
Davide Cozzolino, Giovanni Poggi, and Luisa Verdoliva.
\newblock Recasting residual-based local descriptors as convolutional neural
  networks: an application to image forgery detection.
\newblock In {\em Proceedings of the 5th ACM Workshop on Information Hiding and
  Multimedia Security}, pages 159--164, 2017.

\bibitem{dang2020detection}
Hao Dang, Feng Liu, Joel Stehouwer, Xiaoming Liu, and Anil~K Jain.
\newblock On the detection of digital face manipulation.
\newblock In {\em Proceedings of the IEEE/CVF Conference on Computer Vision and
  Pattern Recognition}, pages 5781--5790, 2020.

\bibitem{DeepFakes}
DeepFakes.
\newblock Deepfakes github.
\newblock \url{http://github.com/deepfakes/faceswap}, 2017.
\newblock Accessed 2020-08-18.

\bibitem{du2019towards}
Mengnan Du, Shiva Pentyala, Yuening Li, and Xia Hu.
\newblock Towards generalizable forgery detection with locality-aware
  autoencoder.
\newblock {\em arXiv preprint arXiv:1909.05999}, 2019.

\bibitem{durall2020watch}
Ricard Durall, Margret Keuper, and Janis Keuper.
\newblock Watch your up-convolution: Cnn based generative deep neural networks
  are failing to reproduce spectral distributions.
\newblock In {\em Proceedings of the IEEE/CVF Conference on Computer Vision and
  Pattern Recognition}, pages 7890--7899, 2020.

\bibitem{durall2019unmasking}
Ricard Durall, Margret Keuper, Franz-Josef Pfreundt, and Janis Keuper.
\newblock Unmasking deepfakes with simple features.
\newblock {\em arXiv preprint arXiv:1911.00686}, 2019.

\bibitem{FaceSwap}
FaceSwap.
\newblock Faceswap github.
\newblock \url{https://github.com/MarekKowalski/FaceSwap}, 2016.
\newblock Accessed 2020-08-18.

\bibitem{fridrich2012rich}
Jessica Fridrich and Jan Kodovsky.
\newblock Rich models for steganalysis of digital images.
\newblock {\em IEEE Transactions on Information Forensics and Security},
  7(3):868--882, 2012.

\bibitem{goodfellow2014generative}
Ian Goodfellow, Jean Pouget-Abadie, Mehdi Mirza, Bing Xu, David Warde-Farley,
  Sherjil Ozair, Aaron Courville, and Yoshua Bengio.
\newblock Generative adversarial nets.
\newblock In {\em Advances in neural information processing systems}, pages
  2672--2680, 2014.

\bibitem{he2016deep}
Kaiming He, Xiangyu Zhang, Shaoqing Ren, and Jian Sun.
\newblock Deep residual learning for image recognition.
\newblock In {\em Proceedings of the IEEE conference on computer vision and
  pattern recognition}, pages 770--778, 2016.

\bibitem{karras2019style}
Tero Karras, Samuli Laine, and Timo Aila.
\newblock A style-based generator architecture for generative adversarial
  networks.
\newblock In {\em Proceedings of the IEEE conference on computer vision and
  pattern recognition}, pages 4401--4410, 2019.

\bibitem{kim2012face}
Gahyun Kim, Sungmin Eum, Jae~Kyu Suhr, Dong~Ik Kim, Kang~Ryoung Park, and
  Jaihie Kim.
\newblock Face liveness detection based on texture and frequency analyses.
\newblock In {\em 2012 5th IAPR international conference on biometrics (ICB)},
  pages 67--72. IEEE, 2012.

\bibitem{kingma2014adam}
Diederik~P Kingma and Jimmy Ba.
\newblock Adam: A method for stochastic optimization.
\newblock {\em arXiv preprint arXiv:1412.6980}, 2014.

\bibitem{korshunov2018deepfakes}
Pavel Korshunov and S{\'e}bastien Marcel.
\newblock Deepfakes: a new threat to face recognition? assessment and
  detection.
\newblock {\em arXiv preprint arXiv:1812.08685}, 2018.

\bibitem{korshunova2017fast}
Iryna Korshunova, Wenzhe Shi, Joni Dambre, and Lucas Theis.
\newblock Fast face-swap using convolutional neural networks.
\newblock In {\em Proceedings of the IEEE International Conference on Computer
  Vision}, pages 3677--3685, 2017.

\bibitem{li2020face}
Lingzhi Li, Jianmin Bao, Ting Zhang, Hao Yang, Dong Chen, Fang Wen, and Baining
  Guo.
\newblock Face x-ray for more general face forgery detection.
\newblock In {\em Proceedings of the IEEE/CVF Conference on Computer Vision and
  Pattern Recognition}, pages 5001--5010, 2020.

\bibitem{li2020sharp}
Xiaodan Li, Yining Lang, Yuefeng Chen, Xiaofeng Mao, Yuan He, Shuhui Wang, Hui
  Xue, and Quan Lu.
\newblock Sharp multiple instance learning for deepfake video detection.
\newblock In {\em Proceedings of the 28th ACM International Conference on
  Multimedia}, pages 1864--1872, 2020.

\bibitem{li2018ictu}
Yuezun Li, Ming-Ching Chang, and Siwei Lyu.
\newblock In ictu oculi: Exposing ai created fake videos by detecting eye
  blinking.
\newblock In {\em 2018 IEEE International Workshop on Information Forensics and
  Security (WIFS)}, pages 1--7. IEEE, 2018.

\bibitem{li2018exposing}
Yuezun Li and Siwei Lyu.
\newblock Exposing deepfake videos by detecting face warping artifacts.
\newblock {\em arXiv preprint arXiv:1811.00656}, 2018.

\bibitem{li2020celeb}
Yuezun Li, Xin Yang, Pu Sun, Honggang Qi, and Siwei Lyu.
\newblock Celeb-df: A large-scale challenging dataset for deepfake forensics.
\newblock In {\em Proceedings of the IEEE/CVF Conference on Computer Vision and
  Pattern Recognition}, pages 3207--3216, 2020.

\bibitem{maaten2008visualizing}
Laurens van~der Maaten and Geoffrey Hinton.
\newblock Visualizing data using t-sne.
\newblock {\em Journal of machine learning research}, 9(Nov):2579--2605, 2008.

\bibitem{masi2020two}
Iacopo Masi, Aditya Killekar, Royston~Marian Mascarenhas, Shenoy~Pratik
  Gurudatt, and Wael AbdAlmageed.
\newblock Two-branch recurrent network for isolating deepfakes in videos.
\newblock {\em arXiv preprint arXiv:2008.03412}, 2020.

\bibitem{matern2019exploiting}
Falko Matern, Christian Riess, and Marc Stamminger.
\newblock Exploiting visual artifacts to expose deepfakes and face
  manipulations.
\newblock In {\em 2019 IEEE Winter Applications of Computer Vision Workshops
  (WACVW)}, pages 83--92. IEEE, 2019.

\bibitem{nguyen2019multi}
Huy~H Nguyen, Fuming Fang, Junichi Yamagishi, and Isao Echizen.
\newblock Multi-task learning for detecting and segmenting manipulated facial
  images and videos.
\newblock {\em arXiv preprint arXiv:1906.06876}, 2019.

\bibitem{nguyen2019capsule}
Huy~H Nguyen, Junichi Yamagishi, and Isao Echizen.
\newblock Capsule-forensics: Using capsule networks to detect forged images and
  videos.
\newblock In {\em ICASSP 2019-2019 IEEE International Conference on Acoustics,
  Speech and Signal Processing (ICASSP)}, pages 2307--2311. IEEE, 2019.

\bibitem{nirkin2019fsgan}
Yuval Nirkin, Yosi Keller, and Tal Hassner.
\newblock Fsgan: Subject agnostic face swapping and reenactment.
\newblock In {\em Proceedings of the IEEE international conference on computer
  vision}, pages 7184--7193, 2019.

\bibitem{petrov2020deepfacelab}
Ivan Petrov, Daiheng Gao, Nikolay Chervoniy, Kunlin Liu, Sugasa Marangonda,
  Chris Um{\'e}, Jian Jiang, Luis RP, Sheng Zhang, Pingyu Wu, et~al.
\newblock Deepfacelab: A simple, flexible and extensible face swapping
  framework.
\newblock {\em arXiv preprint arXiv:2005.05535}, 2020.

\bibitem{pu2016variational}
Yunchen Pu, Zhe Gan, Ricardo Henao, Xin Yuan, Chunyuan Li, Andrew Stevens, and
  Lawrence Carin.
\newblock Variational autoencoder for deep learning of images, labels and
  captions.
\newblock In {\em Advances in neural information processing systems}, pages
  2352--2360, 2016.

\bibitem{10.1007/978-3-030-58610-2_6}
Yuyang Qian, Guojun Yin, Lu Sheng, Zixuan Chen, and Jing Shao.
\newblock Thinking in frequency: Face forgery detection by mining
  frequency-aware clues.
\newblock In Andrea Vedaldi, Horst Bischof, Thomas Brox, and Jan-Michael Frahm,
  editors, {\em Computer Vision -- ECCV 2020}, pages 86--103, Cham, 2020.
  Springer International Publishing.

\bibitem{rahmouni2017distinguishing}
Nicolas Rahmouni, Vincent Nozick, Junichi Yamagishi, and Isao Echizen.
\newblock Distinguishing computer graphics from natural images using
  convolution neural networks.
\newblock In {\em 2017 IEEE Workshop on Information Forensics and Security
  (WIFS)}, pages 1--6. IEEE, 2017.

\bibitem{rossler2019faceforensics++}
Andreas Rossler, Davide Cozzolino, Luisa Verdoliva, Christian Riess, Justus
  Thies, and Matthias Nie{\ss}ner.
\newblock Faceforensics++: Learning to detect manipulated facial images.
\newblock In {\em Proceedings of the IEEE International Conference on Computer
  Vision}, pages 1--11, 2019.

\bibitem{sabir2019recurrent}
Ekraam Sabir, Jiaxin Cheng, Ayush Jaiswal, Wael AbdAlmageed, Iacopo Masi, and
  Prem Natarajan.
\newblock Recurrent convolutional strategies for face manipulation detection in
  videos.
\newblock {\em Interfaces (GUI)}, 3(1), 2019.

\bibitem{sabour2017dynamic}
Sara Sabour, Nicholas Frosst, and Geoffrey~E Hinton.
\newblock Dynamic routing between capsules.
\newblock In {\em Advances in neural information processing systems}, pages
  3856--3866, 2017.

\bibitem{selvaraju2017grad}
Ramprasaath~R Selvaraju, Michael Cogswell, Abhishek Das, Ramakrishna Vedantam,
  Devi Parikh, and Dhruv Batra.
\newblock Grad-cam: Visual explanations from deep networks via gradient-based
  localization.
\newblock In {\em Proceedings of the IEEE international conference on computer
  vision}, pages 618--626, 2017.

\bibitem{stuchi2017improving}
Jos{\'e}~A Stuchi, Marcus~A Angeloni, Rodrigo~F Pereira, Levy Boccato,
  Guilherme Folego, Paulo~VS Prado, and Romis~RF Attux.
\newblock Improving image classification with frequency domain layers for
  feature extraction.
\newblock In {\em 2017 IEEE 27th International Workshop on Machine Learning for
  Signal Processing (MLSP)}, pages 1--6. IEEE, 2017.

\bibitem{thies2019deferred}
Justus Thies, Michael Zollh{\"o}fer, and Matthias Nie{\ss}ner.
\newblock Deferred neural rendering: Image synthesis using neural textures.
\newblock {\em ACM Transactions on Graphics (TOG)}, 38(4):1--12, 2019.

\bibitem{thies2016face2face}
Justus Thies, Michael Zollhofer, Marc Stamminger, Christian Theobalt, and
  Matthias Nie{\ss}ner.
\newblock Face2face: Real-time face capture and reenactment of rgb videos.
\newblock In {\em Proceedings of the IEEE conference on computer vision and
  pattern recognition}, pages 2387--2395, 2016.

\bibitem{wang2020high}
Haohan Wang, Xindi Wu, Zeyi Huang, and Eric~P Xing.
\newblock High-frequency component helps explain the generalization of
  convolutional neural networks.
\newblock In {\em Proceedings of the IEEE/CVF Conference on Computer Vision and
  Pattern Recognition}, pages 8684--8694, 2020.

\bibitem{wang2020cnn}
Sheng-Yu Wang, Oliver Wang, Richard Zhang, Andrew Owens, and Alexei~A Efros.
\newblock Cnn-generated images are surprisingly easy to spot... for now.
\newblock In {\em Proceedings of the IEEE/CVF Conference on Computer Vision and
  Pattern Recognition}, pages 8695--8704, 2020.

\bibitem{xuan2019generalization}
Xinsheng Xuan, Bo Peng, Wei Wang, and Jing Dong.
\newblock On the generalization of gan image forensics.
\newblock In {\em Chinese Conference on Biometric Recognition}, pages 134--141.
  Springer, 2019.

\bibitem{xue2020faster}
Shengke Xue, Wenyuan Qiu, Fan Liu, and Xinyu Jin.
\newblock Faster image super-resolution by improved frequency-domain neural
  networks.
\newblock {\em Signal, Image and Video Processing}, 14(2):257--265, 2020.

\bibitem{yang2019exposing}
Xin Yang, Yuezun Li, and Siwei Lyu.
\newblock Exposing deep fakes using inconsistent head poses.
\newblock In {\em ICASSP 2019-2019 IEEE International Conference on Acoustics,
  Speech and Signal Processing (ICASSP)}, pages 8261--8265. IEEE, 2019.

\bibitem{zhou2017two}
Peng Zhou, Xintong Han, Vlad~I Morariu, and Larry~S Davis.
\newblock Two-stream neural networks for tampered face detection.
\newblock In {\em 2017 IEEE Conference on Computer Vision and Pattern
  Recognition Workshops (CVPRW)}, pages 1831--1839. IEEE, 2017.

\end{thebibliography}
}

\end{document}